\documentclass{article}
\usepackage[preprint]{neurips_2026}
\workshoptitle{Linguistic Principles for Foundation Models (LP4FM)}

\usepackage{float}
\usepackage{placeins}   % for \FloatBarrier
\usepackage[utf8]{inputenc}
\usepackage[T1]{fontenc}
\usepackage{hyperref}
\usepackage{url}
\usepackage{booktabs}
\usepackage{graphicx}
\usepackage{amsmath}
\usepackage{amsfonts}
\usepackage{enumitem}
\usepackage{nicefrac}
\usepackage{microtype}
\usepackage{xcolor}

\newcommand{\cfg}[1]{\texttt{#1}}
\newcommand{\ind}{\mathbf{1}}

\newcommand{\actual}{\textsc{actual}}
\newcommand{\random}{\textsc{random}}
\newcommand{\human}{\textsc{human}}

\title{Instruction-Tuned Models Locally Produce More Syntactic Overlap -- But Not More Conditional Reuse -- Than Humans Do}

\author{%
  Zandi Eberstadt \\
  University of Oxford \\
  \texttt{zandi.eberstadt@cs.ox.ac.uk}
}

\begin{document}
\maketitle
\begin{abstract}
Syntactic convergence — the tendency of speakers to adapt in language towards the grammatical profiles of their interlocutors — is a well-documented, largely implicit feature of human dialogue. Whether large language models exhibit analogous syntactic convergence towards human users relative to human baselines (including across a broad range of syntactic constructions) remains an open question. Using substitution-paradigm data from Blevins et al. [2026] in which model generations replace one speaker's turns in pre-existing human dialogues, this study measures turn-adjacent reuse of context-free grammar (CFG) rules  across sixteen open-weight Llama and Gemma models (1B–70B, pretrained and instruction-tuned) at 1,901 matched positions per model. Every model showed greater CFG-rule overlap with the preceding human turn than with a sampled unrelated human prime (a context-associated effect that does not isolate syntactic priming from topical continuity), and this actual-versus-random difference was larger for lower-frequency rules. Each instruction-tuned model also showed greater natural-output overlap with the actual prime than the human response it replaced, and all eight matched architecture pairs exhibited greater actual-prime overlap after instruction tuning. However, relative to pretrained variants, instruction-tuned outputs overlapped more with unrelated primes, showed a smaller actual-versus-random increment, and had lower conditional rule-reuse odds once target rule-set size was held constant. In exploratory analyses, each model exhibited greater mean lexical and semantic similarity to the preceding turn than matched human responses did. Instruction-tuned models additionally produced responses with greater mean semantic similarity than their pretrained counterparts in all eight architecture pairs, whereas the lexical similarity results were more heterogeneous.

\end{abstract} 

\section{Introduction}

When a speaker mirrors the grammatical constructions of her interlocutor, she engages in a process called syntactic convergence. For instance, after hearing ``I gave you the book,'' she may be more likely to respond with ``you sent your mother the letter'' (reusing the double-object dative construction) than with the equally grammatical ``you sent the letter to your mother'' \citep{bock1986syntactic,branigan2000}. Syntactic convergence is widely considered to be relatively impervious to conscious awareness and is often explained through structural priming mechanisms \citep{bock1986syntactic,pickering2004}.

As entities capable of producing syntactically coherent text, large language models (LLMs) are now routine interlocutors at population scale. Symmetrical LLM syntactic alignment \citep{kandra2025, https://doi.org/10.1111/cogs.70106, chen2026} and structural priming \citep{sinclair2022structural, michaelov-etal-2023-structural, jumelet2024, cai2024large} have been studied before, as has stylometric accommodation of LLMs \citep{blevins2026eacl} (including towards humans \citep{blevins2026bothways}). However,  whether and how LLMs syntactically converge directionally towards humans -- relative to matched human responses and across a broad inventory of syntactic constructions -- remains, to date, unanswered.

Analyzing data from \citet{blevins2026eacl}'s substitution paradigm, I use a measurement pipeline for directional local syntactic convergence in human-LLM dialogue and apply it to dyadic conversations from sixteen open-weight models spanning two families, parameter counts ranging from 1B to 70B, with both pretrained and instruction-tuned counterparts. Specifically, I measure whether LLM-generated turns reuse the context-free grammar (CFG) rules of the human turns to which they are responding, both relative to an unrelated-prime baseline and relative to the original human responses that the LLM output replaces.

This comparison yields four principal results. First, all sixteen
models reuse actual-prime CFG rules above an unrelated-prime baseline
(RQ1). Second, every instruction-tuned model shows greater
natural-output syntactic alignment than the matched human response (although decomposition by unrelated-prime overlap and target structural
opportunity qualifies a simple account of stronger copying under
instruction tuning) (RQ2). 
Third, rule frequency appears to shape actual-prime reuse differently across
humans, pretrained models, and instruction-tuned models (RQ3). Finally, each of the sixteen models shows greater mean lexical and semantic similarity to the preceding turn than matched human responses do (RQ4).

It is worth being precise about the terminology used in this paper. I use \emph{convergence} (which may have both long-term and short-term instantiations -- priming being an example of the latter) to refer specifically to accommodation towards another interlocutor, and \emph{alignment} to describe the resulting state or degree of similarity. Importantly, greater rule overlap with an actual prime can arise because a response reuses syntax more strongly, because it contains more structural material and therefore affords more opportunities for overlap, or both. I use \emph{total natural-output alignment} for the amount of actual-prime overlap in responses as generated, \emph{actual-versus-random increment} for the added overlap attributable to the true conversational prime relative to an unrelated one, and \emph{conditional reuse propensity} for analyses that hold target rule-set size constant. 

\section{Background}

While communicative mirroring occurs across a number of behaviors (both linguistically and paralinguistically), syntactic convergence is a theoretically distinctive convergence dimension. Grammatical form can recur independently of lexical content \citep{bock1986syntactic}. Historically, a dominant experimental paradigm for measuring syntactic \textit{priming} has involved manipulating pre-specified constructions and coding, in a binary fashion, whether a participant subsequently produced the same form \citep{bock1986syntactic, branigan2000}. Context-free grammar (CFG)-based corpus methods instead represent utterances as production rules extracted from constituency parse trees, permitting broad-coverage measurement without pre-specifying constructions for the analysis \citep{reitter2014}.

To measure LLM linguistic accommodation, \citet{blevins2026eacl} introduce a substitution paradigm that replaces one speaker's turns in pre-existing human–human dialogues, permitting direct comparison between an LLM-generated response and what a human speaker actually said at the identical conversational position. Their stylometric analyses (including of token novelty and utterance length) show that model and human convergence profiles differ by feature and tuning regime. By contrast, \cite{kandra2025} examine syntactic structure, restrict their analysis to \textit{intra}-LLM conversations, and use a symmetrical aggregate measure. Other experiments have provided complementary evidence of human-like structural priming signatures in language models including inverse-frequency effects \citep{jumelet2024}, but have not compared generated dialogue responses against matched human respondents across a range of syntactic constructions. The need to understand LLM (syntactic) accommodation is increasingly pressing,  as \cite{augustin2026characterizing} propose an "amplification spiral" in which linguistic alignment, hyperpersonalized generation, and sycophancy may co-occur in the construction of delusional user beliefs. 

The present study, then, asks four questions:
\begin{enumerate}[
    label=\textbf{RQ\arabic*:},
    leftmargin=*,
    itemsep=1pt,
    topsep=2pt,
    parsep=0pt
]
    \item Do LLMs reuse the CFG rules of the immediately
preceding human turn above an unrelated-prime baseline?

    \item Do LLMs reuse CFG rules more or less
than a matched human respondent would at the same turn position?

    \item Does reuse vary with rule
frequency in the same way across human, pretrained, and instruction-tuned responses?

    \item Do human–model differences in similarity to the preceding turn appear at the lexical and semantic levels as well as the syntactic one?

\end{enumerate}

\section{Methods}

\subsection{Data and substitution paradigm}

Human turns are drawn from DailyDialog \citep{li2017}, a dataset of dyadic English conversations, along with the substituted versions released by \citet{blevins2026eacl} in which LLM-generated completions replace one speaker's utterances from turn six onward. Following their preprocessing, conversations were restricted to two-speaker dialogues with at least six turns, yielding 707 conversations; their model generations and preprocessing code are publicly released. The full set of 707 conversations was analyzed for every model. 

Sixteen open-weight language models are evaluated: the Llama-3.1 (8B, 70B) and Llama-3.2 (1B, 3B) families \citep{grattafiori2024llama} and the Gemma-3 family (1B, 4B, 12B, 27B) \citep{gemmateam2025gemma3technicalreport}, each in both its pretrained and instruction-tuned variant. Because every model completes the identical conversations and because the human turns are held fixed across all versions, responses are directly comparable across models and against the original human responses in the same conversational contexts. Generated responses were capped at 40 new tokens. From the second substituted position onward, a model's generation is conditioned in part on its own earlier generations; the immediately preceding turn itself is nevertheless always human in the analyzed prime--target pair.

Ultimately, there were 1{,}918 substituted positions per model. Seventeen positions had no eligible rule in the immediately adjacent human prime and were excluded, leaving 1{,}901 exactly matched positions per model. No non-adjacent prime was substituted for a missing one. Human and model targets with no eligible rules were retained and contributed zero outcomes for every eligible prime rule.

\subsection{Syntactic annotation and rule eligibility}

All utterances, both human and LLM-generated, were parsed using benepar's \cfg{benepar\_en3\_large} constituency parser \citep{kitaev2018,kitaev2019}, with spaCy \citep{montani2023explosion} used for tokenization and segmentation. Prior to parsing, utterance text was normalized by collapsing whitespace sequences to a single space.

A constituency parse represents an utterance as a tree whose
nonterminal nodes encode syntactic categories such as S, NP, and VP. For example, a parse of ``The reader reviewed the paper''
contains the productions
\[
\mathrm{S}\rightarrow\mathrm{NP}\ \mathrm{VP},\qquad
\mathrm{NP}\rightarrow\mathrm{DT}\ \mathrm{NN},\qquad
\mathrm{VP}\rightarrow\mathrm{VBD}\ \mathrm{NP}.
\]

A CFG production records how one nonterminal node expands into its immediate children. Preterminal part-of-speech nodes expand directly to terminal words, as in \texttt{DT $\rightarrow$ The}.
For an utterance $u$, let $\rho(u)$ denote the multiset of CFG
productions extracted from its sentence-level parse trees. 
$R(u)$ is the set of distinct eligible rule types appearing in
$\rho(u)$, where a rule is eligible if it is non-unary and occurs more
than once in the original human corpus. Thus, repeated occurrences of
the same rule within an utterance contribute only once. Frequencies were calculated once from human turns only, preventing model outputs from changing rule eligibility or the frequency scale.  (Lexical preterminal productions are unary and are therefore already excluded). The human corpus contained 10{,}887 rule types in total, of which 1{,}405 met the global eligibility criterion; 1{,}138 appeared in the analyzed prime sets.

\subsection{Matched actual, human, and unrelated-prime conditions}

For matched position $i$, write $p_i$ for the prime (the human turn at $t-1$), $h_i$ for the original human response at $t$, $m_i$ for the model generation at $t$, and $\tilde p_i$ for a human prime sampled with a fixed seed from a different conversation. One $\tilde p_i$ was drawn per position and shared across all sixteen models (to ensure that model comparisons do not inherit different random baselines). The three conditions pair $(p_i,h_i)$ in the \human{} condition, $(p_i,m_i)$ in the \actual{} model condition, and $(\tilde p_i,m_i)$ in the \random{} condition. Thus, the human and model targets respond to the same potential source of syntactic priming, while the unrelated-prime condition holds the model response fixed and varies only the prime.

For each condition $\kappa$ at position $i$ and each eligible prime rule $r_j$, the binary outcome is
\begin{equation}
 y_{ij}^{\kappa}=\ind\!\left[r_j\in R(\mathrm{target}_i^{\kappa})\right].
 \label{eq:outcome}
\end{equation}
Rule frequency was represented as mean-centered $\log F_H(r_j)$, and prime structural size as mean-centered $\log |R(\mathrm{prime}_i)|$. No additive constant was required before either logarithm: eligible rules satisfy $F_H(r)>1$, and positions with empty prime rule sets were excluded before analysis. In the structural-opportunity analyses, target size was represented as mean-centered $\log(|R(\mathrm{target}_i)|+1)$. (The $+1$ offset is required only for target size because a valid human or model response may contain no eligible rules). This procedure produced 39{,}222 rule-level observations per model, or 627{,}552 observations across the sixteen per-model datasets.
\subsection{Statistical models}

For each model separately, a logistic mixed-effects model was fitted over
condition
$\kappa\in\{\textsc{human},\textsc{actual},\textsc{random}\}$.
Let
$x_j=\widetilde{\log F_H(r_j)}$
denote the mean-centered log frequency of rule $r_j$, and let
$q_i^\kappa=\widetilde{\log|R(p_i^\kappa)|}$
denote the mean-centered log size of the relevant prime rule set.
Here, $p_i^\kappa=p_i$ in the \textsc{human} and \textsc{actual}
conditions, whereas $p_i^\kappa=\tilde p_i$ in the
\textsc{random} condition.

Using \textsc{human} as the reference condition, the linear predictor
was

\begin{align}
\operatorname{logit}
\Pr\!\left(y_{ij}^{\kappa}=1\right)
={}&
\beta_0
+\beta_{\mathrm{A}}\,
 \mathbf{1}\!\left\{\kappa=\textsc{actual}\right\}
+\beta_{\mathrm{R}}\,
 \mathbf{1}\!\left\{\kappa=\textsc{random}\right\}
\nonumber\\
&+\beta_f x_j
+\beta_{\mathrm{A}f}\,
 \mathbf{1}\!\left\{\kappa=\textsc{actual}\right\}x_j
+\beta_{\mathrm{R}f}\,
 \mathbf{1}\!\left\{\kappa=\textsc{random}\right\}x_j
\nonumber\\
&+\beta_q q_i^\kappa
+b_i+c_j ,
\label{eq:permodel}
\end{align}

where
$b_i\sim\mathcal{N}(0,\sigma_{\mathrm{position}}^2)$
is a random intercept for matched position and
$c_j\sim\mathcal{N}(0,\sigma_{\mathrm{rule}}^2)$
is a random intercept for rule.

Under this parameterization,
$\beta_{\mathrm{A}}$ is the actual-model-versus-human contrast at
centered mean rule frequency and prime size, and
$\beta_{\mathrm{A}f}$ is the corresponding difference in the frequency
slope. The actual-versus-random contrast is

\begin{equation}
\beta_{\mathrm{A-R}}
=
\beta_{\mathrm{A}}-\beta_{\mathrm{R}},
\label{eq:actual-random}
\end{equation}

and the interaction testing whether that increment varies with rule
frequency is

\begin{equation}
\beta_{\mathrm{A-R}\times f}
=
\beta_{\mathrm{A}f}-\beta_{\mathrm{R}f}.
\label{eq:actual-random-frequency}
\end{equation}

For reporting, the same specification was also fit with
\textsc{random} as the reference condition, which yields the
actual-versus-random contrast and its standard error directly.
Because condition interacts with frequency, each condition coefficient
is evaluated at centered mean rule frequency.

A per-model sensitivity fit added the centered target-size term
$t^{\kappa}_i$ as a main effect, yielding the conditional contrast
$\beta_{A|t}$. This fit asks how condition differences change when
the amount of eligible syntactic structure in the target is held
constant (and since target size is produced during generation, these
estimates are interpreted as descriptive conditional comparisons
rather than causal adjustments).

Comparisons between sixteen separately estimated coefficients do not
themselves constitute a formal test of instruction tuning. I therefore
grouped the models into eight matched architecture pairs and fit two
pooled GLMMs over model-target rows. The actual-prime model
($N=212{,}544$) included architecture pair, tuning, centered log rule
frequency, centered log prime size, the tuning-by-frequency interaction,
and random intercepts for position, pair-by-position, and rule. This
model estimates the instruction-minus-pretraining difference in total
actual-prime overlap within the evaluated model suite.

A second pooled model used the actual- and unrelated-prime model rows
($N=415{,}008$) and included the full
tuning-by-prime-type-by-frequency interaction. Its
tuning-by-actual-prime coefficient tests whether instruction tuning
changes the actual-versus-random increment, while the three-way
interaction tests whether that change varies with rule frequency. As a
complementary architecture-level consistency check, I calculated the raw
actual-prime reuse difference within each matched pair and applied an
exact two-sided sign test.

To separate total natural-output alignment from conditional reuse
propensity, I fit a pooled actual-prime model containing the
tuning-by-target-size interaction. The full model used all
212{,}544 rule rows. Because pretrained and instruction-tuned
target-size distributions differed substantially, I also fit the same
specification within a central common-support sample. For each
architecture pair, positions were retained when their log target size
lay in the intersection of the pretrained and instruction-tuned
5th--95th percentile ranges, yielding 149{,}101 rule rows. Contrasts at
the 10th, 50th, and 90th percentiles of the supported target-size
distribution were calculated as linear combinations of the tuning and
tuning-by-target-size coefficients.

All models were fit in R using \cfg{lme4::glmer()}
\citep{bates2015}, with binomial errors, the \cfg{bobyqa} optimizer,
200{,}000 maximum function evaluations, and
$\mathrm{nAGQ}=1$. All fits were nonsingular with no convergence warnings (Table~\ref{tab:diagnostics}).

\subsection{Lexical and semantic similarity}

Beyond the primary inferential analysis, two exploratory analyses were conducted to better characterize the observed convergence behavior. Lexical similarity was measured as TF--IDF cosine similarity using one vocabulary shared across all primes, human targets, and model targets (6{,}481 terms). Semantic similarity was measured as the cosine between L2-normalized \cfg{all-MiniLM-L6-v2} sentence embeddings \citep{reimers2019}. For each matched position, I calculated human-minus-model similarity (so negative values indicate that the model response is closer to the prime). For each model and metric, the mean difference was estimated with an intercept-only OLS regression and conversation-clustered standard errors. Holm correction \citep{holm1979} was applied separately across the sixteen lexical tests and the sixteen semantic tests. Seventy-eight blank model outputs were excluded, leaving 30{,}338 valid model-position comparisons.

\section{Results}

\subsection{Primary result I: all sixteen models exceed the
unrelated-prime baseline (RQ1)}

Every model reused actual-prime rules more than rules from its unrelated
prime. The actual-versus-random coefficient ranged from
$\beta=.706$ to $1.453$, with all
$p\le6.5\times10^{-31}$ (Table ~\ref{tab:permodel}).
Thus, all sixteen evaluated models showed greater syntactic reuse from
the true preceding human turn than from an unrelated human turn.

\subsection{Primary result II: instruction-tuned models show greater
natural-output alignment than matched humans, while pretrained results
are mixed (RQ2)}

\begin{figure}[t]
  \centering
  \includegraphics[width=\linewidth]{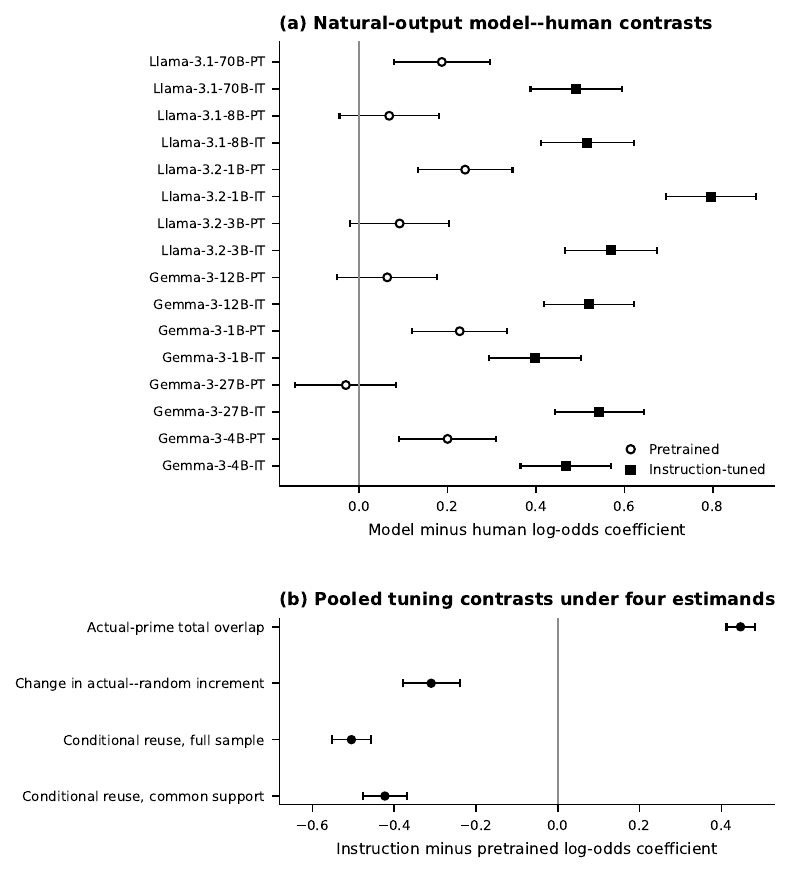}
  \caption{\textbf{Primary and pooled syntactic results.} (a) Per-model actual-prime contrasts against the matched human response at centered mean rule frequency; error bars are 95\% Wald intervals. (b) Pooled instruction-minus-pretraining coefficients under four estimands. Positive values favor instruction tuning. Conditional coefficients are evaluated at the target-size centering mean. PT = pretrained; IT = instruction-tuned.}
  \label{fig:main}
\end{figure}

Under the human-reference primary model, every instruction-tuned model reused actual-prime rules more than the matched human response at centered mean frequency: $\beta=.398$--$.796$, all $p\le8.3\times10^{-14}$. The largest coefficient belonged to Llama-3.2-1B-Instruct ($\beta=.796$). Scale effects were not monotonic across families, however, so these data do not warrant a more general claim that natural-output syntactic alignment varies with parameter count (Figure~\ref{fig:main}a).

Pretrained models were closer to the human baseline as well as more heterogeneous; their coefficients ranged from $-.030$ to $.240$, and four were significantly positive while four were not distinguishable from zero. All eight matched architecture pairs showed greater actual-prime reuse
under instruction tuning (Table ~\ref{tab:pair-rates}), with raw instruction-minus-pretraining
differences of $.022$--$.094$ and an exact two-sided sign-test
$p=.0078$. The pooled rule-level GLMM estimated a corresponding
within-suite instruction-minus-pretraining contrast of
$\beta=.447$, 95\% CI $[.412,.481]$, and
$\mathrm{OR}=1.56$.

\paragraph{Decomposing the instruction-tuning difference.}

Nonetheless, the actual-versus-random pooled model complicates a simple account in which instruction tuning straightforwardly strengthens CFG reuse from the immediately preceding turn. Under the unrelated-prime reference, instruction-tuned outputs already had higher overlap than pretrained outputs ($\beta=.743$). The pretrained actual-versus-random increment was positive ($\beta=1.177$), but the tuning-by-actual-prime interaction was negative ($\beta=-.310$, 95\% CI $[-.379,-.240]$, $\mathrm{OR}=.734$, $p=2.8\times10^{-18}$). Thus, instruction-tuned models produced more total overlap under the actual prime while receiving a smaller incremental boost from replacing the unrelated prime with the true conversational prime.

The three-way tuning-by-prime-type-by-frequency interaction was positive ($\beta=.083$, $p=2.8\times10^{-6}$). Both tuning regimes retained a low-frequency-weighted actual-prime effect, but that concentration was weaker under instruction tuning. Pretrained and instruction-tuned responses then appear to differ in \emph{how much} syntax overlaps between the prime and its subsequent generated response, but also in how much of that overlap is specifically attributable to the true conversational prime.

\begin{table}[t]
\centering
\small
\caption{Pooled instruction-tuning effects under distinct estimands. Conditional coefficients are evaluated at the target-size centering mean; percentile-specific contrasts appear in Table ~\ref{tab:target}.}
\label{tab:pooled}
\begin{tabular}{p{.43\linewidth}rrrr}
\toprule
Estimand & $\beta$ & OR & 95\% CI & $p$ \\
\midrule
Actual-prime total overlap & .447 & 1.563 & [.412, .481] & $4.3\!\times\!10^{-144}$ \\
Change in actual--random increment & $-.310$ & .734 & [$-.379$, $-.240$] & $2.8\!\times\!10^{-18}$ \\
Conditional reuse, full sample & $-.505$ & .604 & [$-.552$, $-.458$] & $3.0\!\times\!10^{-98}$ \\
Conditional reuse, common support & $-.423$ & .655 & [$-.477$, $-.369$] & $1.5\!\times\!10^{-53}$ \\
\bottomrule
\end{tabular}
\end{table}

\paragraph{Conditioning on target structural opportunity.}

Additionally, target rule-set size was a predictor of overlap. In the full pooled sensitivity model, a one-unit increase in centered log target size multiplied the odds of rule overlap by $7.29$ for the pretrained reference condition ($\beta=1.987$). Once target size was conditioned upon, the instruction-tuning coefficient reversed: $\beta=-.505$, 95\% CI $[-.552,-.458]$, $\mathrm{OR}=.604$, $p=3.0\times10^{-98}$ (Table~\ref{tab:pooled}). The same direction held in the common-support sample ($\beta=-.423$,
95\% CI $[-.477,-.369]$, OR $=.655$, $p=1.5\times10^{-53}$), though
retention in that sample ranged from 33.2\% to 95.5\% across
pair-by-tuning cells, so it should not
be read as representative of every natural instruction-tuned response. Moreover, instruction-tuned variants showed lower conditional reuse at the 10th, 50th, and 90th percentiles of the supported target-size distribution in both analyses (Table~\ref{tab:target}). 

\paragraph{The human comparison under the same conditioning.}
The per-model sensitivity fits (Table~\ref{tab:permodel}, column
$\beta_{A-H|t}$) apply the same conditioning to the model--human contrast,
and they reorder it. Unconditionally, pretrained models sat at or near the
human baseline ($\beta_{A-H}=-.030$ to $.240$) while every instruction-tuned
model exceeded it. Holding target rule-set size constant reverses this pattern, in that
every pretrained model exceeded the matched human response by
$\beta_{A-H|t}=.683$ to $.922$ (all $p<.001$; OR $=1.98$ to $2.51$),
whereas instruction-tuned models exceeded it more modestly
($\beta_{A-H|t}=.166$ to $.525$ in five architectures) or not detectably
at all (Gemma-3-12B-IT, $.069$; Gemma-3-27B-IT, $-.081$; Gemma-3-4B-IT,
$-.023$). The elevated natural-output alignment of instruction-tuned
responses therefore rests substantially on the amount of eligible structure
those responses contain, while the strongest conditional reuse propensity in the
suite belongs to pretrained models.

\subsection{Primary result III: frequency profiles differ across
humans, pretrained models, and instruction-tuned models (RQ3)}

Across all sixteen models, the actual-versus-random interaction with
frequency was negative ($\beta=-.504$ to $-.231$, all
$p\le9.7\times10^{-12}$): the additional reuse attributable to the true
conversational prime rather than an unrelated one was larger for rarer
rules.

Relative to matched human responses, all eight pretrained
model-by-frequency interactions were negative, seven significantly so
($\beta=-.190$ to $-.052$). The instruction-tuned interactions were
more varied: four were significantly positive, one was significantly
negative, and three were nonsignificant ($\beta=-.088$ to $.121$).

In the pooled actual-prime model, tuning interacted positively with
frequency ($\beta=.153$, $p=3.2\times10^{-64}$), indicating that the
instruction-tuning advantage in total actual-prime overlap was larger
for more frequent rules. Humans, pretrained models, and instruction-tuned
models therefore did not exhibit a single shared frequency profile.

\subsection{Exploratory finding: models show greater mean lexical and semantic similarity than matched human responses (RQ4)}

All sixteen models were more lexically similar to the prime than the matched human response (Table~\ref{tab:lexical}). Human-minus-model TF--IDF differences ranged from $-.083$ to $-.036$, and all sixteen tests remained significant after Holm correction. Likewise, all sixteen models were more semantically similar to the prime, with differences from $-.104$ to $-.018$ and all Holm-adjusted tests significant. The instruction-tuned variant was more semantically similar than its matched pretrained counterpart in all eight architecture pairs. The lexical effect was not consistent, however; four pairs moved towards greater lexical similarity under tuning and four moved in the opposite direction. 

\section{Discussion}

This study measured directional turn-adjacent syntactic convergence in human-LLM dialogue, analyzing \citet{blevins2026eacl}'s substitution-paradigm corpus and drawing upon a CFG-based measurement framework; findings are discussed below.

First, every instruction-tuned model showed greater natural-output syntactic alignment than the matched human response, and all eight matched architecture pairs showed greater total actual-prime overlap under tuning. A plausible interpretation is that instruction tuning optimizes models for local "responsiveness" -- engaging directly with the immediately preceding turn through partial reuse of its phrasing -- which would be expected to yield alignment with the prime across numerous linguistic levels at once. Additionally (and beyond the syntactic domain), humans have been found to evaluate more positively others who are perceived to behave similarly to themselves \citep{montoya2008actual}, which raises the possibility that preference-based post-training could (either directly or indirectly) reward certain forms of linguistic similarity. The present analysis, however, does not identify the training mechanism responsible for the observed pattern.

Second (and complicating a takeaway that "instruction-tuned models converge more than pretrained ones"), instruction-tuned outputs overlap more with \textit{unrelated} primes, their actual-versus-random increment is smaller, and their tuning advantage reverses after target structure is held constant. Put otherwise, instruction-tuned models produce responses containing more eligible syntactic structure, which affords more opportunities for overlap, while pretrained outputs exhibit higher conditional rule-reuse odds at comparable levels of eligible target structure. This framing also reconciles the present results with {\citet{blevins2026eacl}}, who find that instruction-tuned models generally converge less than pretrained variants on several stylometric features. They suggest that reduced convergence may in part reflect a tendency to bring novel material into exchanges during instruction tuning; such a tendency could also extend to syntax, although this interpretation does remain speculative.

Third, at the lexical and semantic levels, all sixteen models exceeded human similarity to the preceding turn. In every pair, the instruction-tuned model was more semantically similar than its pretrained counterpart, while the lexical direction split. Notably, convergence in human dialogue is a graded and socially-modulated signal \citep{giles1991accommodation}. In these models, similarity appears to be biased towards the immediately preceding turn, though future analyses could examine differential model convergence (to various speakers and across registers) in more depth. To be fair, I would resist reading the similarity measures as convergence proper: prime-to-response similarity conflates accommodation with topical echoing, particularly under a 40-token generation cap. These exploratory lexical and semantic analyses do not include an unrelated-prime baseline and therefore cannot independently establish accommodation beyond topical continuity.

Finally, for every model, the incremental CFG reuse was larger among the lower-frequency rules. Interestingly, human structural priming effects are also strongest for structures that are
rare or dispreferred \citep{scheepers2003syntactic, bernolet2010does, jaeger2013alignment}. That this pattern appears in every model, including the smallest pretrained variants, suggests that sensitivity to the local syntactic environment is present before dialogue-specific instruction tuning. Rare CFG rules may also be more lexically or topically specific, however, so this frequency profile should not be taken as decisive evidence of a syntax-specific priming mechanism. In humans, this frequency profile has been taken as evidence for error-driven implicit learning \citep{chang2006becoming, jaeger2013alignment}; that account arguably cannot transfer directly  to this paradigm, since no weights are updated during generation and any analogue must operate within the context window. A prediction-based account is nonetheless available in principle, in that predicting text in context plausibly rewards representations in which recently encountered structures are more available, though the present data cannot adjudicate between mechanistic accounts. 

Two implications follow from the analyses. First, with respect to understanding LLM psycholinguistics, these findings supply a measure of how closely a generated turn tracks the human turn it answers, while suggesting that alignment measured at one level (e.g., the syntactic) does not straightforwardly license inference about another (e.g., the lexical). Second, these findings contribute to emerging discussions of linguistic alignment in deployed systems and its potential risks \citep{boyd2026artificial, augustin2026characterizing}. The present paradigm operationalizes one local linguistic component
that may be relevant to broader accounts of this phenomenon.

\section{Scope, Limitations, and Extensions}

Certain limitations qualify the present study. First, all human-human results are derived from DailyDialog, a corpus of short dialogues crawled from English-learning websites. Obtaining data and conducting analyses on other registers would be instructive to determine if similar patterns generalize. The 40-token generation cap together with the brevity of DailyDialog turns  bounds the amount of structure available for reuse (human responses are not subject to the same generation cap, making response structure an especially important part of the estimand). Relatedly, the present study is restricted to English; future researchers are encouraged to apply the substitution paradigm to multilingual corpora to address whether these convergence profiles are stable across typologically diverse languages.

Importantly, the actual-versus-random comparison does not isolate syntax-specific priming. Unlike the unrelated prime, the actual prime shares discourse context, topic, and potentially lexical material with the target response. Consequently, greater CFG-rule overlap under the actual prime may partly reflect topical continuity or lexically associated constructions rather than syntactic accommodation independently of content. Though the matched human comparisons provide a useful reference within the same conversational positions, they do not eliminate this limitation. The results are therefore best understood as evidence of local context-associated syntactic overlap. Future work could employ lexically-controlled primes or experimental manipulations of syntactic structure.

A further limitation concerns the dependence structure of the inferential models. Although the primary GLMMs include random intercepts for matched position and CFG rule, multiple analyzed positions originate from the same conversation, and later model responses may depend on earlier model-generated turns. These dependencies are not fully represented by the specified random-effects structure. Accordingly, the reported confidence intervals and significance tests should be interpreted conditional on that specification.  Results are also conditional on a single unrelated-prime draw per position and on the benepar parser, whose accuracy on conversational and model-generated text was not independently audited.
 
\section{Conclusion}

Across sixteen models, turn-adjacent CFG overlap exceeded an unrelated-prime baseline; instruction tuning increased total overlap but lowered conditional reuse once target structure was held constant. Against the matched human baseline, all sixteen exceeded human lexical and semantic similarity to the preceding turn, and every instruction-tuned model additionally exceeded human syntactic alignment in its natural output. 

Whether the patterns observed here reflect the tuning distribution, preference optimization, or a broader property of next-token prediction under dialogue formatting is an empirical inquiry left to future work. More generally speaking, these results raise questions related to human-centered computing that extend beyond the present study; it, in any event, remains to be seen whether and how LLM syntactic convergence affects task efficacy and user wellbeing. \FloatBarrier
\label{mainend}
\typeout{MAIN_TEXT_LAST_PAGE=\thepage}
\clearpage

\bibliographystyle{plainnat}
\bibliography{references}

\clearpage
\appendix

\section{Per-model syntactic estimates}
\label{app:permodel}

Table~\ref{tab:permodel} reports the key per-model coefficients. The
$\beta_{A-H|t}$ column is a per-model sensitivity coefficient and should not be
interpreted as the pooled tuning contrast reported in Table~\ref{tab:pooled}.

\begin{table}[H]
\centering
\caption{Per-model GLMM coefficients. $\beta_{A-H}$ is the primary
model-versus-human contrast and $\beta_{A-R}$ the actual-versus-random
contrast, both at centered mean rule frequency and prime size.
$\beta_{(A-H)\times f}$ is the model--human frequency interaction and
$\beta_{(A-R)\times f}$ the model--random frequency interaction; the two differ
because the former varies the respondent at a fixed prime while the latter
varies the prime at a fixed target. Negative $\beta_{(A-R)\times f}$ indicates
that the actual-prime increment is larger for lower-frequency rules.
$\beta_{A-H|t}$   is the model--human contrast from the sensitivity fit that
additionally conditions on centered target rule-set size~$t$ (distinct from prime size $q_i^\kappa$ in
Equation~\ref{eq:permodel}). Stars denote two-sided Wald tests.}
\label{tab:permodel}
\small
\setlength{\tabcolsep}{5pt}
\begin{tabular}{lccccc}
\toprule
Model & $\beta_{A-H}$ & $\beta_{A-R}$ & $\beta_{(A-H)\times f}$
      & $\beta_{(A-R)\times f}$ & $\beta_{A-H|t}$ \\
\midrule
Llama-3.1-70B-PT       &      0.187$^{***}$ &      1.453$^{***}$ &   $-$0.190$^{***}$ &   $-$0.504$^{***}$ &      0.922$^{***}$ \\
Llama-3.1-70B-IT       &      0.491$^{***}$ &      0.860$^{***}$ &              0.013 &   $-$0.301$^{***}$ &       0.166$^{**}$ \\
Llama-3.1-8B-PT        &              0.068 &      1.218$^{***}$ &           $-$0.052 &   $-$0.419$^{***}$ &      0.693$^{***}$ \\
Llama-3.1-8B-IT        &      0.516$^{***}$ &      0.954$^{***}$ &       0.087$^{**}$ &   $-$0.304$^{***}$ &      0.213$^{***}$ \\
Llama-3.2-1B-PT        &      0.240$^{***}$ &      1.368$^{***}$ &   $-$0.163$^{***}$ &   $-$0.472$^{***}$ &      0.847$^{***}$ \\
Llama-3.2-1B-IT        &      0.796$^{***}$ &      1.041$^{***}$ &      0.121$^{***}$ &   $-$0.340$^{***}$ &      0.258$^{***}$ \\
Llama-3.2-3B-PT        &              0.091 &      1.148$^{***}$ &     $-$0.071$^{*}$ &   $-$0.355$^{***}$ &      0.704$^{***}$ \\
Llama-3.2-3B-IT        &      0.570$^{***}$ &      0.822$^{***}$ &      0.105$^{***}$ &   $-$0.231$^{***}$ &      0.198$^{***}$ \\
Gemma-3-12B-PT         &              0.063 &      1.107$^{***}$ &   $-$0.119$^{***}$ &   $-$0.352$^{***}$ &      0.755$^{***}$ \\
Gemma-3-12B-IT         &      0.520$^{***}$ &      0.790$^{***}$ &           $-$0.026 &   $-$0.308$^{***}$ &              0.069 \\
Gemma-3-1B-PT          &      0.227$^{***}$ &      1.317$^{***}$ &   $-$0.131$^{***}$ &   $-$0.420$^{***}$ &      0.838$^{***}$ \\
Gemma-3-1B-IT          &      0.398$^{***}$ &      1.149$^{***}$ &    $-$0.088$^{**}$ &   $-$0.411$^{***}$ &      0.525$^{***}$ \\
Gemma-3-27B-PT         &           $-$0.030 &      1.146$^{***}$ &   $-$0.102$^{***}$ &   $-$0.392$^{***}$ &      0.683$^{***}$ \\
Gemma-3-27B-IT         &      0.543$^{***}$ &      0.706$^{***}$ &       0.084$^{**}$ &   $-$0.254$^{***}$ &           $-$0.081 \\
Gemma-3-4B-PT          &      0.200$^{***}$ &      1.416$^{***}$ &   $-$0.160$^{***}$ &   $-$0.492$^{***}$ &      0.895$^{***}$ \\
Gemma-3-4B-IT          &      0.467$^{***}$ &      0.775$^{***}$ &              0.010 &   $-$0.301$^{***}$ &           $-$0.023 \\
\bottomrule
\end{tabular}
\vspace{2pt}
\\{\footnotesize $^{*}p<.05$, $^{**}p<.01$, $^{***}p<.001$.}
\end{table}

\section{Matched-pair raw rates}

\begin{table}[h]
\centering
\small
\caption{Raw actual-prime reuse rates for the eight matched architecture pairs. All differences favor instruction tuning; exact two-sided sign-test $p=.0078$. Differences are computed from unrounded rates and may differ from the displayed subtraction by .001.}
\begin{tabular}{lrrr}
\toprule
Architecture pair & Pretrained & Instruction & Difference \\
\midrule
Gemma-3-12B & 0.155 & 0.207 & +0.052 \\
Gemma-3-1B & 0.167 & 0.189 & +0.022 \\
Gemma-3-27B & 0.144 & 0.226 & +0.082 \\
Gemma-3-4B & 0.160 & 0.205 & +0.044 \\
Llama-3.1-70B & 0.159 & 0.208 & +0.048 \\
Llama-3.1-8B & 0.163 & 0.217 & +0.054 \\
Llama-3.2-1B & 0.163 & 0.257 & +0.094 \\
Llama-3.2-3B & 0.158 & 0.224 & +0.067 \\
\bottomrule
\label{tab:pair-rates}
\end{tabular}
\end{table}

\section{Target-size contrasts and common support}

\begin{table}[h]
\centering
\small
\caption{Instruction-versus-pretraining contrasts at supported target sizes. Negative values indicate lower conditional reuse under instruction tuning. The median contrasts differ slightly from Table~\ref{tab:pooled} because the empirical median is close to, but not exactly, the centering mean.}
\begin{tabular}{llrrrr}
\toprule
Analysis & Target size & $\beta$ & OR & 95\% CI & $p$ \\
\midrule
Full sample & 10th percentile & -0.357 & 0.699 & [-0.437, -0.278] & $9.0\times10^{-19}$ \\
Full sample & Median & -0.504 & 0.604 & [-0.551, -0.457] & $6.3\times10^{-98}$ \\
Full sample & 90th percentile & -0.633 & 0.531 & [-0.674, -0.592] & $3.0\times10^{-200}$ \\
Common support & 10th percentile & -0.286 & 0.751 & [-0.390, -0.183] & $6.2\times10^{-8}$ \\
Common support & Median & -0.422 & 0.655 & [-0.476, -0.369] & $2.2\times10^{-53}$ \\
Common support & 90th percentile & -0.542 & 0.582 & [-0.601, -0.482] & $5.0\times10^{-71}$ \\
\bottomrule
\label{tab:target}
\end{tabular}
\end{table}

\begin{table}[h]
\centering
\scriptsize
\caption{Position retention in the central common-support analysis. The 22,049 retained positions are 72.5\% of 30,416 position-by-model observations; they contribute 149,101 rule rows, 70.2\% of the full 212,544 rows, because retained positions contain slightly fewer eligible prime rules on average.}
\begin{tabular}{llrrr}
\toprule
Architecture & Tuning & Retained & Total & Retained \% \\
\midrule
Gemma-3-12B & Pretrained & 1816 & 1901 & 95.5 \\
Gemma-3-1B & Pretrained & 1809 & 1901 & 95.2 \\
Gemma-3-27B & Pretrained & 1215 & 1901 & 63.9 \\
Gemma-3-4B & Pretrained & 1506 & 1901 & 79.2 \\
Llama-3.1-70B & Pretrained & 1805 & 1901 & 95.0 \\
Llama-3.1-8B & Pretrained & 1570 & 1901 & 82.6 \\
Llama-3.2-1B & Pretrained & 1318 & 1901 & 69.3 \\
Llama-3.2-3B & Pretrained & 1576 & 1901 & 82.9 \\
Gemma-3-12B & Instruction & 1122 & 1901 & 59.0 \\
Gemma-3-1B & Instruction & 1698 & 1901 & 89.3 \\
Gemma-3-27B & Instruction & 631 & 1901 & 33.2 \\
Gemma-3-4B & Instruction & 1112 & 1901 & 58.5 \\
Llama-3.1-70B & Instruction & 1178 & 1901 & 62.0 \\
Llama-3.1-8B & Instruction & 1344 & 1901 & 70.7 \\
Llama-3.2-1B & Instruction & 1088 & 1901 & 57.2 \\
Llama-3.2-3B & Instruction & 1261 & 1901 & 66.3 \\
\bottomrule
\label{tab:positionretention}
\end{tabular}
\end{table}

\clearpage
\section{Lexical and semantic similarity}

\begin{figure}[H]
\centering
\includegraphics[width=.9\linewidth]{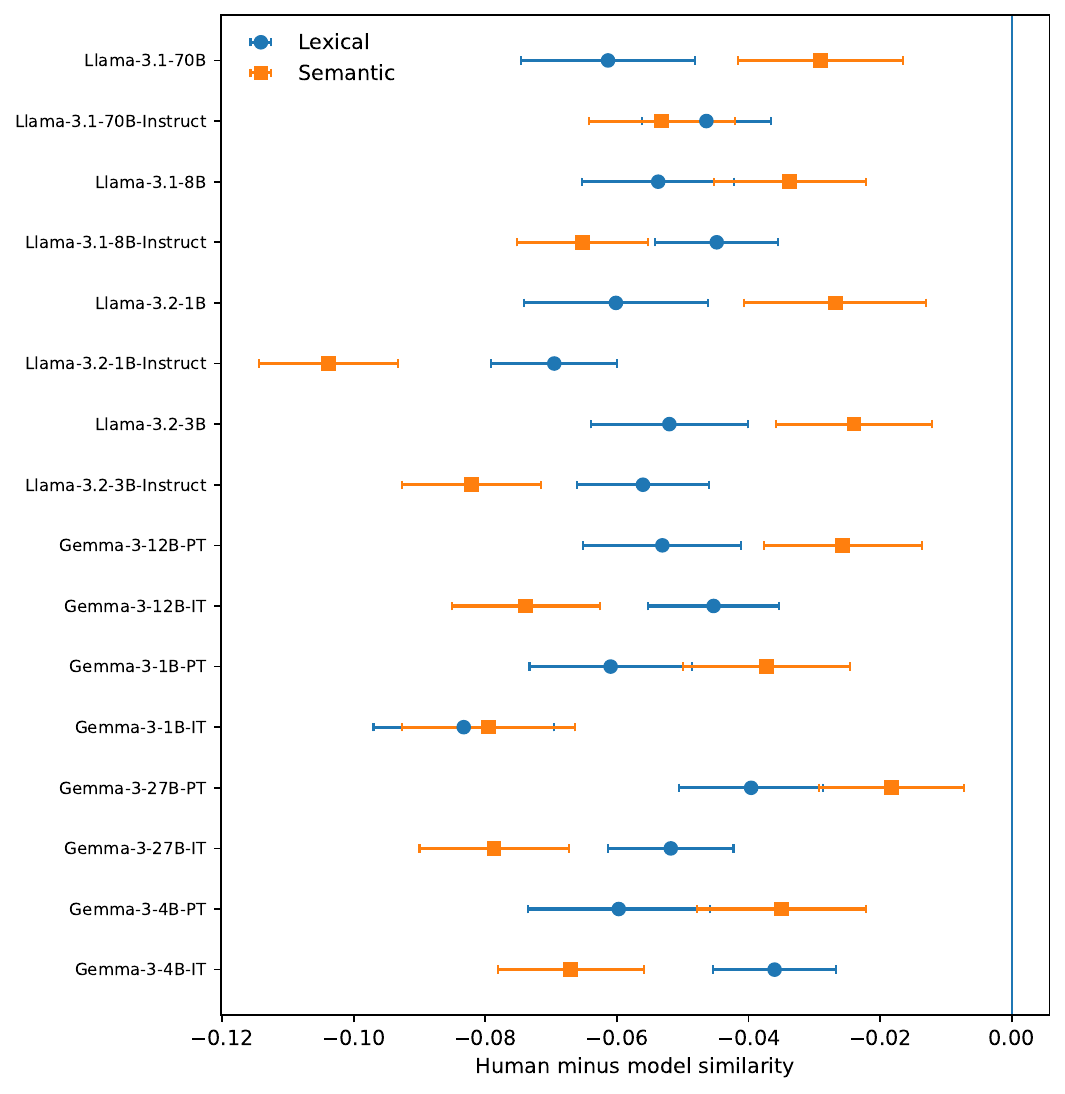}
\caption{Human-minus-model similarity differences. Negative values mean that the model response is closer to the prime. Error bars are 95\% conversation-clustered intervals. PT = pretrained; IT = instruction-tuned.}
\label{fig:lexical}
\end{figure}

\begin{table}[H]
\centering
\scriptsize
\setlength{\tabcolsep}{3.5pt}
\caption{Human-minus-model similarity differences and Holm-adjusted $p$-values.}
\begin{tabular}{lrrrr}
\toprule
Model & Lexical diff. & Lexical $p_{\mathrm{Holm}}$ & Semantic diff. & Semantic $p_{\mathrm{Holm}}$ \\
\midrule
Llama-3.1-70B & -0.061 & $7.2\times10^{-19}$ & -0.029 & $2.5\times10^{-5}$ \\
Llama-3.1-70B-Instruct & -0.046 & $1.4\times10^{-19}$ & -0.053 & $4.9\times10^{-20}$ \\
Llama-3.1-8B & -0.054 & $7.5\times10^{-19}$ & -0.034 & $7.6\times10^{-8}$ \\
Llama-3.1-8B-Instruct & -0.045 & $4.7\times10^{-20}$ & -0.065 & $1.9\times10^{-36}$ \\
Llama-3.2-1B & -0.060 & $1.1\times10^{-16}$ & -0.027 & $2.8\times10^{-4}$ \\
Llama-3.2-1B-Instruct & -0.070 & $1.3\times10^{-44}$ & -0.104 & $1.4\times10^{-81}$ \\
Llama-3.2-3B & -0.052 & $6.3\times10^{-17}$ & -0.024 & $2.2\times10^{-4}$ \\
Llama-3.2-3B-Instruct & -0.056 & $1.7\times10^{-26}$ & -0.082 & $8.9\times10^{-52}$ \\
Gemma-3-12B-PT & -0.053 & $2.4\times10^{-17}$ & -0.026 & $1.1\times10^{-4}$ \\
Gemma-3-12B-IT & -0.045 & $3.6\times10^{-18}$ & -0.074 & $9.0\times10^{-37}$ \\
Gemma-3-1B-PT & -0.061 & $3.6\times10^{-21}$ & -0.037 & $6.6\times10^{-8}$ \\
Gemma-3-1B-IT & -0.083 & $1.6\times10^{-31}$ & -0.080 & $1.5\times10^{-31}$ \\
Gemma-3-27B-PT & -0.040 & $1.1\times10^{-12}$ & -0.018 & $1.1\times10^{-3}$ \\
Gemma-3-27B-IT & -0.052 & $2.1\times10^{-25}$ & -0.079 & $6.2\times10^{-41}$ \\
Gemma-3-4B-PT & -0.060 & $1.1\times10^{-16}$ & -0.035 & $5.4\times10^{-7}$ \\
Gemma-3-4B-IT & -0.036 & $6.8\times10^{-14}$ & -0.067 & $2.2\times10^{-31}$ \\
\bottomrule
\label{tab:lexical}
\end{tabular}
\end{table}

\section{Diagnostics}

\begin{table}[h]
\centering
\small
\caption{Diagnostics for the pooled and target-size models. All fits
were nonsingular and produced no convergence messages.}
\label{tab:diagnostics}
\begin{tabular}{lrrr}
\toprule
Analysis & Observations & Singular & Max.\ absolute gradient \\
\midrule
Actual-prime pooled tuning
& 212{,}544 & No & .013 \\
Actual-versus-random pooled gap
& 415{,}008 & No & .056 \\
Target-size full sample
& 212{,}544 & No & .021 \\
Target-size common support
& 149{,}101 & No & .007 \\
\bottomrule
\end{tabular}
\end{table}
The analysis used a fixed random-prime seed of 42. Preliminary exploratory fits used $\mathrm{nAGQ}=0$; all coefficients reported in this manuscript were refit with $\mathrm{nAGQ}=1$. 

\end{document}